\documentclass[twocolumn,twoside]{IEEEtran}
\usepackage{amsmath,amssymb,amsthm,wasysym,epsfig,color,graphicx,epstopdf,url,bm,nicefrac,tcolorbox}
\usepackage{xcolor,xpatch}

\usepackage{subcaption}

\usepackage{threeparttable}
\usepackage{enumerate}
\usepackage{caption}
\usepackage{amsmath,graphicx,bm,accents}
\usepackage{booktabs}

\usepackage{multicol}
\usepackage{multirow}

\usepackage{algorithm}
\usepackage{algorithmic}

\allowdisplaybreaks

\makeatletter
\ExplSyntaxOn
\cs_new:Npn \bibColoredItems #1#2
{
	\clist_map_inline:nn {#2} { \cs_new:cpn {bib@colored@##1} {#1} } 
}
\ExplSyntaxOff

\newcommand\bib@setcolor[1]{%
	\ifcsname bib@colored@#1\endcsname
	\expanded{\noexpand\color{\csname bib@colored@#1\endcsname}}%
	\else
	\normalcolor
	\fi
}

\xpatchcmd\@bibitem
{\item}
{\bib@setcolor{#1}\item}
{}{\fail}

\xpatchcmd\@lbibitem
{\item}
{\bib@setcolor{#2}\item}
{}{\fail}
\makeatother

\usepackage{accents}
\makeatletter
\def\wid{\check{{\cc@style\underline{\mskip9.5mu}}}}
\def\Wideubar{\underaccent{{\cc@style\underline{\mskip6mu}}}}
\makeatother

\makeatletter
\def\wideubar{\underaccent{{\cc@style\underline{\mskip9.5mu}}}}
\def\Wideubar{\underaccent{{\cc@style\underline{\mskip6mu}}}}
\makeatother

\makeatletter
\def\widebar{\accentset{{\cc@style\underline{\mskip9.5mu}}}}
\def\Widebar{\accentset{{\cc@style\underline{\mskip6mu}}}}
\makeatother

\theoremstyle{remark}

\newcommand{\gang}[1]{{\color{black}{#1}}}

\allowdisplaybreaks

\begin{document}
	\title{Flexible Job Shop Scheduling via Dual Attention Network Based Reinforcement Learning}
	
	\author{Runqing Wang, Gang Wang,~\IEEEmembership{Member,~IEEE}, Jian Sun,~\IEEEmembership{Senior Member,~IEEE}, \\Fang Deng,~\IEEEmembership{Senior Member,~IEEE}, and Jie Chen,~\IEEEmembership{Fellow,~IEEE}
		\thanks{The work was supported in part by the National Key R\&D Program of China under Grant 2021YFB1714800, in part by the National Natural Science Foundation of China under Grants 62173034, 61925303, 62025301, 62088101, in part by the CAAI-Huawei MindSpore Open Fund, and in part by the Chongqing Natural Science Foundation under Grant 2021ZX4100027. (\emph{Corresponding author: Jian Sun.})
			
			R. Wang, G. Wang, J. Sun, and F. Deng are with the National Key Lab of Autonomous Intelligent Unmanned Systems, School of Automation, Beijing Institute of Technology, Beijing 100081, China, and the Beijing Institute of Technology Chongqing Innovation Center, Chongqing 401120, China
			(e-mail: rqwang@bit.edu.cn; gangwang@bit.edu.cn; sunjian@bit.edu.cn; dengfang@bit.edu.cn). 
			
			J. Chen is with the Department of Control Science and Engineering,
			Tongji University, Shanghai 201804, China, and the National Key Lab of Autonomous Intelligent Unmanned Systems, Beijing Institute of Technology, Beijing 100081, China (e-mail:
			chenjie@bit.edu.cn).
		}
		
	}
	
\maketitle
\allowdisplaybreaks

\begin{abstract}
	Flexible manufacturing has given rise to complex scheduling problems such as the flexible job shop scheduling problem (FJSP). In FJSP, operations can be processed on multiple machines, leading to intricate relationships between operations and machines. Recent works have employed deep reinforcement learning (DRL) to learn priority dispatching rules (PDRs) for solving FJSP. However, the quality of solutions still has room for improvement relative to that by the exact methods such as OR-Tools. To address this issue, this paper presents a novel end-to-end learning framework that weds the merits of self-attention models for deep feature extraction and DRL for scalable decision-making. The complex relationships between operations and machines are represented precisely and concisely, for which a dual-attention network (DAN) comprising several interconnected operation message attention blocks and machine message attention blocks is proposed. The DAN exploits the complicated relationships to construct production-adaptive operation and machine features to support high-quality decision-making. Experimental results using synthetic data as well as public benchmarks corroborate that the proposed approach outperforms both traditional PDRs and the state-of-the-art DRL method. Moreover, it achieves results comparable to exact methods in certain cases and demonstrates favorable generalization ability to large-scale and real-world unseen FJSP tasks.
	
\end{abstract}

\begin{IEEEkeywords}
	Flexible job-shop scheduling, self-attention mechanism, deep reinforcement learning, graph attention networks
\end{IEEEkeywords}

\section{Introduction}
\label{sec:intro}

Industry 4.0
is transforming the way companies manufacture, improve, and distribute their products, by moving towards rapid, intelligent, and flexible manufacturing, which leads to a fundamental change in production capabilities of enterprises \cite{mao2019opportunities,eng2022}. The flexible job-shop scheduling problem (FJSP) is a classic problem that represents the typical scenarios faced by flexible manufacturing. FJSP has gained increasing attention in diverse fields, such as cyber-physical manufacturing \cite{ding2019defining}, cloud computing \cite{arunarani2019task}, and intelligent transportation \cite{satunin2014multi}. The job-shop scheduling problem (JSP), which is a simpler and easier instance of FJSP, is also a core issue in the manufacturing industry \cite{zhang2019review}. JSP involves a set of jobs and machines, where each job consists of multiple operations that must be processed on a given machine in a pre-defined order. The goal is to generate a processing sequence of operations that achieves a meaningful production goal, such as minimum completion time, lateness/tardiness, or production cost. Compared to JSP, FJSP is a more challenging problem as it allows each operation to be processed on multiple different machines.In addition to the operation sequencing problem in JSP,  the machine assignment problem adds further manufacturing flexibility,
rendering FJSP a more complex topology and a much larger solution space. 

In fact, it has been shown that FJSP is a strongly NP-hard problem \cite{xie2019review}.
The combinatorial nature of FJSP makes it challenging to find (near-)optimal solutions using traditional operation research methods such as constraint programming \cite{meng2020mixed}. These methods have intractable computation costs that increase dramatically with the problem size, making them impractical for large-scale applications \cite{demir2013evaluation}. To strike a balance between the solution quality and the computation cost, research in the field has gradually shifted from traditional heuristic and meta-heuristic methods to intelligent methods such as deep learning \cite{zhang2019review} and particularly deep reinforcement learning (DRL) \cite{mnih2013playing}.

Meta-heuristics, including genetic algorithms \cite{wang2019genetic,rooyani2019efficient}, particle swarm optimization \cite{zhang2009effective}, differential evolution \cite{du2018high}, and artificial bee colony \cite{li2019hybrid}, have been widely employed for scheduling problems, and they often find high-quality solutions by means of a complex solution search procedure. In contrast, rule-based heuristics, such as priority dispatching rules (PDRs), are more practical due to their ease of implementation and high efficiency \cite{haupt1989survey}. PDRs repeatedly select the operation or machine with the highest priority based on some prescribed rules until a complete plan is generated. Nonetheless, designing effective PDRs often requires significant expertise and research effort, and they may only perform well in specific tasks.

DRL methods have emerged as promising approaches to solve JSP and FJSP \cite{wang2021review}, which model the scheduling process as a Markov decision process (MDP). In these methods, a parameterized neural network model is designed to receive information about the production environment as state and output the priority of each feasible scheduling action, such as assigning an operation to a machine, forming an end-to-end learning approach \gang{\cite{wang2021dynamic}}. Through training on a set of production process data, the DRL model learns to adaptively select the best action at a state to maximize the total reward, which is related to the production goal. However, the effectiveness and efficiency of these methods heavily depend on the design of the state representation, which is a challenging task. As the problem size increases, the amount of production information grows significantly. Therefore, representing this information and complex constraints in the scheduling environment in a sufficiently and minimally expressive manner poses key challenges for designing the state representation. However, the trained DRL-based model is expected to be capable of solving problems of different sizes based on a single unifying architecture. Additionally, the structural information of the problem, which comprises diverse sets of constraints, also plays a critical role in making JSP and FJSP challenging. The priority of operations or machines is strongly related to these constraints, making it essential for the model to appropriately express and exploit them.

Several attempts have been made to address these challenges. One approach is to use aggregated features of operations or machines, such as average machine workload, to represent the production state instead of separately representing each operation or machine, to achieve a uniform representation for problems of varying sizes. For example, this approach is used in \cite{luo2021real} to characterize the production status of dynamic partial-no-wait FJSP. They designed 20 extracted features as the state and developed a hierarchical DRL-based framework to learn from these features and choose rules from a set of PDRs for the operation and machine selection independently. Similarly, in \cite{du2022reinforcement}, a similar approach for state representation is used to describe FJSP with crane transportation and setup times, and a non-fully connected deep Q-network was employed for selecting compound rules to solve this problem. Although this state representation approach is generic for various production scenarios, such as FJSP with additional production constraints, and simplifies the depiction of FJSP, it compresses the information about the production state and makes little use of the structural information.

Alternatively, the well-known disjunctive graph \cite{brandimarte1993routing} has been widely used in many studies to represent the state of (F)JSP. This graph-based model represents operations as nodes and uses directed edges between nodes to indicate the processing order of operations, allowing it to describe features of each operation and represent the structural information of (F)JSP. Recent research has shown promising results in leveraging representation learning methods such as graph neural networks (GNNs) and self-attention mechanisms \cite{vaswani2017attention} to handle combinatorial optimization problems with complicated constraints, as these methods have the ability to capture the inherent structure of the problem and are size-agnostic.

Integration of these methods with DRL has been explored in various domains such as vehicle routing \cite{nazari2018reinforcement,kool2018attention}, chip design \cite{mirhoseini2021graph}, and production scheduling \cite{kwon2021matrix}, and has demonstrated success in solving extremely large-scale problems \cite{manchanda2020gcomb}. In the domain of JSP, for example, a learning framework that incorporates GNN and DRL was proposed in \cite{zhang2020learning} for generating the priority of operations at each step of the scheduling process using a sparse graph as the state representation. Similarly, the work of \cite{park2021learning} incorporated dynamic attributes in node features and designed a customized GNN for message passing to learn operation embeddings. In \cite{chen2022deep}, graph embedding techniques were utilized to extract features of the disjunctive graph for downstream decision-making, and an attention-based model was proposed to generate solutions for JSP.

Recent works have extended these solutions to handle FJSP by considering additionally machine features and improving decision-making models to handle both operation sequencing and machine assignment. For instance, the work of \cite{lei2022multi} introduced a two-step decision-making framework that uses a disjunctive graph for learning operation embeddings and a multi-layer perceptron (MLP) for passing machine messages. However, this approach may result in limited exploration during training since the priority of machines is only computed for the selected operation. On the other hand, \cite{song2022flexible} designed a heterogeneous graph that considers operations and machines as heterogeneous nodes and connects each machine to operations that it can process. They proposed a two-stage GNN for learning machine and operation embeddings and a decision-making model for selecting operations and machines simultaneously.

In general, albeit these DRL-based solutions yield better performance than traditional PDRs, some challenging problems still remain to be tackled for further improving the model’s performance and narrowing the gap with exact methods. First, a more precise state representation is required which avoids to incorporate information that are irrelevant to decision-making. For instances, completed operations have no contributions to subsequent decision-making at a specific scheduling step, for they will not affect the production status. However, existing methods consider all operations at each step, which may affect the performance and also reduce the efficiency \cite{huang2019convolutional}. Second, diverse relationships between operations and machines require to be represented and exploited more rationally. Although existing methods have considered the operation-operation or operation-machine relationships, they have not modeled the relationship between machines yet. The relationship between machines can be viewed as the competition for remaining unscheduled operations which is crucial for discriminating high-priority machines. 

In response to these challenges, we propose a novel end-to-end learning framework for standard FJSP aiming to minimize the completion time. First, a tight state representation is introduced in the MDP formulation, which incorporates relevant information about decision-relevant operations and machines. Second, a dual attention network, comprising several operation and machine message attention blocks, is proposed to express the complicated relationships of operations and machines concisely. The operation message attention blocks learn to extract operation features by simply exploiting their precedence constraints, and the machine message attention blocks leverage the competitive relationships between machines to extract machine features.
Finally, a size-agnostic decision-making framework is designed to simultaneously address the two critical subproblems in FJSP: operation sequencing and machine assignment. Experiment results show that our approach outperforms traditional PDRs and the state-of-the-art DRL method by a considerable margin on two synthetic data with different distributions. The performance is even comparable to exact methods (with time limits) on some tasks. Moreover, it exhibits favorable generalization ability, as models trained on small-scale instances achieve high-quality solutions when directly applied to solve large-scale or out-of-distribution instances.

The contributions of this work are summarized as follows.
\begin{itemize}
	\item A tight state representation for describing operations and machines in FJSP that is minimal and sufficient for downstream decision-making with the state space decreasing as scheduling proceeds;
	\item A dual-attention network consisting of several operation and machine message attention blocks for deep feature extraction of operations and machines; and,
	\item A size-agnostic DRL-based approach with (markedly) improved performance and generalization capability compared to conventional PDRs and the state-of-the-art DRL method.
\end{itemize}

The remainder of the paper is structured as below. Section \ref{sec:preli} introduces the necessary basics and the FJSP. Section \ref{sec:RDAN} provides the detailed formulation of the proposed method. Section \ref{sec:experi} reports the experimental results with Section \ref{sec:concl} concluding the paper.

\section{Preliminaries and Problem Formulation}
\label{sec:preli}

\subsection{Flexible job-shop scheduling problem and disjunctive graph}

The FJSP can be formally stated as follows. Consider a set of $n$ jobs and $m$ machines,
represented by $\mathcal{J}$ and $\mathcal{M}$, respectively. We assume that all jobs arrive simultaneously at system production time $T_s=0$. Each job $J_i \in \mathcal{J}$ consists of $n_i$ operations which must be assembled in a specific order (i.e. precedence constraints) described by $\mathcal{O}_i = \{O_{i1},O_{i2},\ldots,O_{in_i}\}$. The set of all operations across all jobs is denoted by $\mathcal{O}=\bigcup_{i}\mathcal{O}_i$. Each operation $O_{ij}$ can be 
processable by multiple machines, but can only be processed on a single machine from the set of available and compatible machines 
$\mathcal{M}_{ij} \subseteq  \mathcal{M}$. The associated processing time on machine $M_k \in \mathcal{M}_{ij}$ is given by $p_{ij}^{k}>0$. 
FJSP seeks to design a schedule that determines for each operation an appropriate processing machine and the start time while respecting the following constraints: c1) the operations of $J_i$ must be processed in the order in $\mathcal{O}_i$ (i.e. precedence constraints); c2) each operation must be assigned to  exactly one compatible machine; and c3) each machine can process at most an operation at a time. The goal of FJSP is to minimize the maximum completion time of all jobs; that is, minimizing the total makespan. 
\begin{figure}[t]
	\begin{subfigure}{\linewidth}
		\centering
		\includegraphics[width=0.8\linewidth]{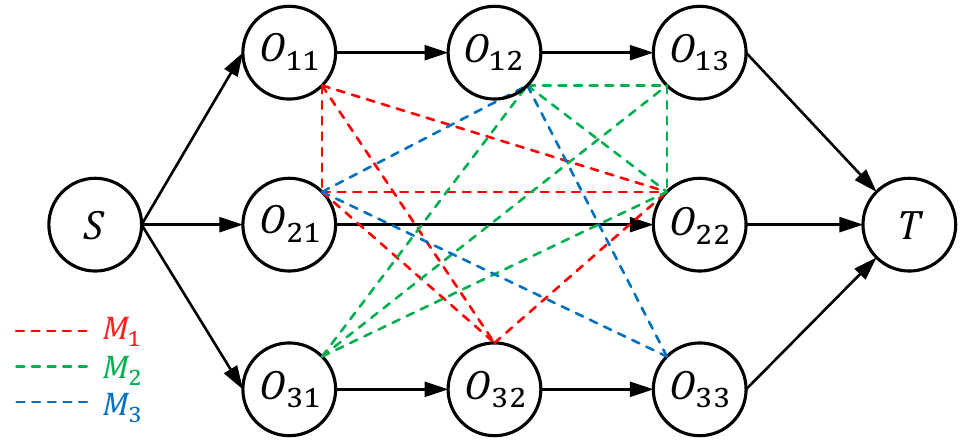}
		\caption{An FJSP instance}
		\label{dg1}
	\end{subfigure}
	\qquad
	\begin{subfigure}{\linewidth}
		\centering
		\includegraphics[width=0.8\linewidth]{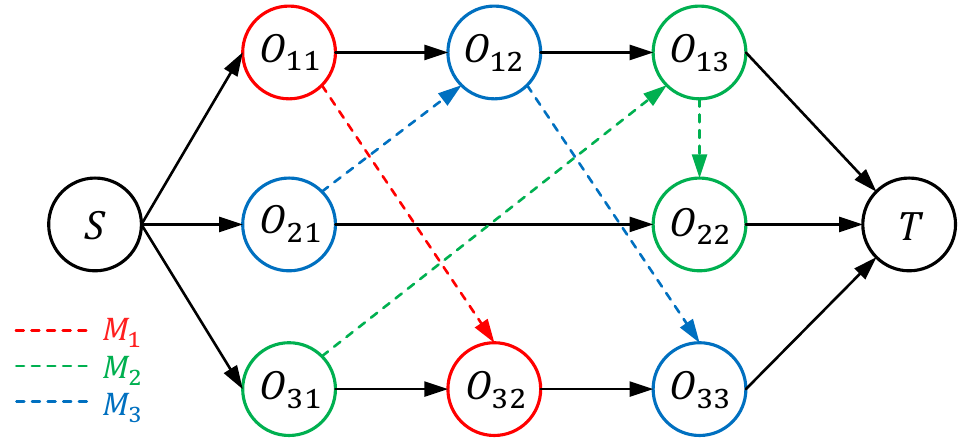}
		\caption{A possible solution}
		\label{dg2}
	\end{subfigure}
	\caption{The disjunctive graph for FJSP.}
	\label{disgraph}
\end{figure}

The disjunctive graph is a well-documented technique for representing the scheduling problems such as JSP and FJSP. In FJSP, a disjunctive graph $G$ can be described by a tuple $(\mathcal{V}=\mathcal{O} \cup \{Start,End\},\mathcal{C},\mathcal{D})$, see Fig. \ref{disgraph}. The node set $\mathcal{V}$ is composed of all operations (i.e., nodes) and two dummy nodes (signifying the start and completion of production). The edge set $\mathcal{C}=\{\langle O_{ij},O_{i,j+1}\rangle| 1 \leq i \leq |J|, 1 \leq j < n_i \}$ comprises all directed edges called conjunctions, which models constraint c1) above; 
and $\mathcal{D}$ is the set of undirected edges or disjunctions, which consists of $m$ groups, denoted by $\mathcal{D}_1,...,\mathcal{D}_m$, respectively. Disjunctions in group $\mathcal{D}_k$ connect operations that machine $M_k$ can process. Consequently, an operation can be linked to disjunctions in different groups according to their flexibility. A solution to FJSP can be obtained by updating these disjunctions. Once an operation $O_{ij}$ is scheduled to be processed on a machine $M_k$, all disjunctions linked to $O_{ij}$ except for the ones in $\mathcal{D}_k$ are then deleted. Meanwhile, disjunctions that link scheduled operations are converted into directed edges, whose direction indicates the processing sequence of operations on corresponding machines. As a result, the disjunctive graph becomes a directed acyclic graph when scheduling is completed.

\subsection{Graph attention network}

Graph attention networks (GATs) \cite{velivckovic2017graph} are one of the most popular and effective architectures in the field of GNNs. The graph attention layer (GAL) is the core block of GATs which uses attention mechanisms to aggregate neighboring information and compute node embeddings. Roughly speaking, a GAL accepts as input a graph with a set $\mathcal{H}=\{\vec{h}_i \in \mathbb{R}^F\}_{i=1}^N $ of nodal feature vectors and outputs for each node a new feature vector $\vec{h}_{i}^{\prime} \in \mathbb{R}^{F^{\prime}}$. 

Letting $\mathcal{N}_i$ collect the first-hop neighbors of node $i$  (including node $i$ itself), a GAL first computes the attention coefficients $e_{i j}$ for each node $i$ and $j \in \mathcal{N}_i$ as follows
\begin{equation}
	\label{eq:gat-1}
	e_{i j}=\text {LeakyReLU}\big(\vec{\mathbf{a}}^\top\big[(\mathbf{W} \vec{h}_i) \| (\mathbf{W} \vec{h}_j)\big]\big)
\end{equation}
where $\mathbf{W}\in\mathbb{R}^{F'\times F}$ is a linear transformation shared by all nodes, the operation $[\cdot \| \cdot ]$ concatenates two vectors to form a larger vector, and $\vec{\mathbf{a}}\in\mathbb{R}^{2F'\times 1}$ is the weight vector of a single-layer neural network with a LeakyReLU($\cdot$) activation function. These attention coefficients are further normalized using the softmax function as follows
\begin{equation}
	\label{eq:gat-2}
	\alpha_{i j}=\operatorname{softmax}_j(e_{i j})=\frac{\exp (e_{i j})}{\sum_{k \in \mathcal{N}_i} \exp (e_{i k})},\quad \forall i.
\end{equation}
Finally, the weighted sum of transformed neighboring features using the normalized attention coefficients is computed, which is then fed into a nonlinear activation function $\sigma:\mathbb{R}\to\mathbb{R}$ (e.g., the exponential linear unit or ELU) to yield a new feature vector
\begin{equation}
	\label{eq:gat-3}
	\vec{h}_i^{\prime}=\sigma\bigg(\sum_{j \in \mathcal{N}_i} \alpha_{i j} \mathbf{W} \vec{h}_j\bigg),\quad \forall i.
\end{equation}
A GAT can have multiple GALs connected one by one forward similar to feedforward neural networks. The node embeddings at  the last layer are 
employed in downstream tasks. 


\section{The Proposed Method}
\label{sec:RDAN}

\begin{figure}[t]
	\centering 
	\includegraphics[scale=0.37]{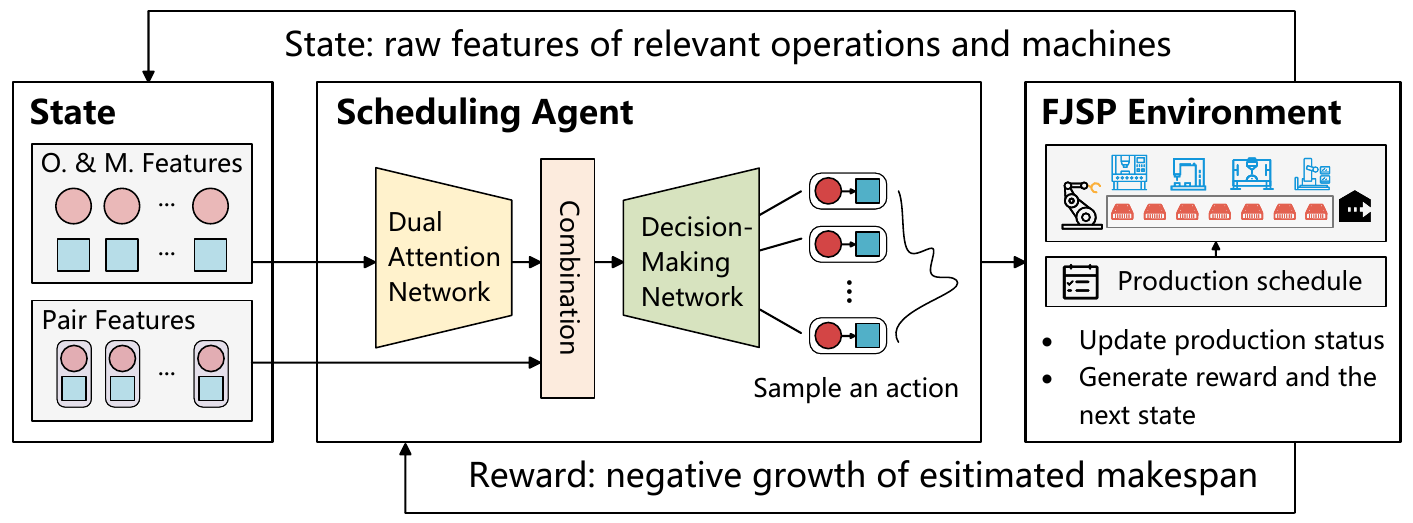}
	\caption{\small{An overview of the proposed method. 
	}}
	\label{fig:workflow}
\end{figure}

In this section, we present our main results including the MDP formulation and a new end-to-end approach for standard FJSP. To this aim, 
we first formulate FJSP as a Markov decision process by defining its states, actions, state transitions, and rewards. Then, we introduce the proposed  learning approach for FJSP, which is termed Dual Attention Network based reInforcEment Learning and DANIEL for short. It builds on GAT-style self-attention mechanisms and deep reinforcement learning. The workflow of the proposed method is depicted in Fig. \ref{fig:workflow}. Specifically, our framework consists of i) a dual attention network for deep feature extraction of and between operations and machines; and, ii) a DRL-based decision-making network that considers operation-selection and machine-selection as a whole and outputs a probability distribution prioritizing available operation-machine pairs. 
At last, we demonstrate how to train the proposed model.

\subsection{MDP formulation of FJSP}

The scheduling process can be understood as dynamically assigning a ready operation to a compatible and idle machine. In this way, the decision point $t$ is the production system time $T_s(t)$ when there is at least a compatible operation-machine pair $(O_{ij},M_k)$ such that $O_{ij}$ can be processed on machine $M_k$ at time $T_s(t)$. At step $t$, the DRL model receives a state $s_t$ from the environment and takes an action $a_t$ that assigns a compatible pair to start processing immediately at time $T_s(t)$, for which the FJSP environment returns a reward $r_t$ related to the makespan. A solution of FJSP can be obtained by repeating this procedure $ | \mathcal{O}  | $ times for the entire set of operations in the task. The MDP is defined as follows.

\emph{State.} We propose a tight state representation that sufficiently characterizes the operations and machines  relevant to decision-making. Specifically, the operations can be categorized into three groups according to their status at the time, namely, completed operations, being processed operations, and unscheduled operations. Since the decision at step $t$ only depends on the production status at system time $T_s(t)$, all operations, except for the completed ones which will not affect subsequent scheduling, are referred to as relevant operations. Likewise, the machines that cannot process any of the remaining unscheduled operations are also termed irrelevant to subsequent scheduling. Therefore, information about irrelevant operations and machines are not meaningful for scheduling and will not be recorded in the state. Let $\mathcal{O}_u(t)$ and $\mathcal{M}_u(t)$ be the set of relevant operations and machines at step $t$, and $\mathcal{A}(t)$ the set of compatible operation-machine pairs. The state $s_t$ is a set of feature vectors, including $h_{O_{ij}} \in \mathbb{R}^{10}$ for each operation $O_{ij}\in \mathcal{O}_u(t)$, $h_{M_k} \in \mathbb{R}^{8}$ for each machine $M_k \in \mathcal{M}_u(t)$, and $h_{(O_{ij},M_k)} \in \mathbb{R}^{8}$ for each compatible operation-machine pair $(O_{ij},M_k) \in \mathcal{A}(t)$. They are carefully handcrafted to provide a minimal yet sufficient description about the static and dynamic properties of scheduling-relevant entities; see the appendix for more details. It is worth stressing that, different from existing works, the state space gets smaller as the production/scheduling proceeds (until it becomes empty meaning that all operations have been scheduled and processing ends), making the representation computationally appealing for real-time and large-scale applications. 
   
\emph{Action.}  The set $\mathcal{A}(t)$ of all compatible operation-machine pairs defines the action space  at step $t$, which becomes smaller as more operations get scheduled for processing across time. We also refer to the operations in $\mathcal{A}(t)$ as the candidates, all of which form the candidate set $J_c(t)$.
 
\emph{State transition.} An action $a_t$ corresponds to the processing of an operation on a machine. Upon taking the action, the environment  (i.e., the production status of all operations and machines) changes. Then, the sets $\mathcal{O}_u(t)$, $\mathcal{M}_u(t)$ and $\mathcal{A}(t)$ as well as features of relevant entities are updated, and a new state $s_{t+1}$ is obtained.

\emph{Reward.} The reward $r_t$ should be designed to guide the agent to choose actions that are helpful in reducing the maximum completion time of all operations in the task, namely the makespan, denoted by $C_\text{max}$. Inspired by the design in \cite{zhang2020learning}, 
we estimate a lower bound of the completion time $\underline{C}(O_{ij}, s_t)$ for each operation $O_{ij}$ at step $t$. If $O_{ij}$ has been scheduled, the value equals to its actual completion time $C_{ij}$ which is accurately known and can be computed. For all unscheduled operations, the lower bound can be approximated by iteratively running the recursion $\underline{C}(O_{ij}, s_t) = \underline{C}(O_{i(j-1)}, s_t)+\min \limits_{k \in M_{ij}} p_{ij}^{k}$ assuming without loss of generality that $\underline{C}(O_{i,0})=0$. The estimated maximum completion time $ \max \limits_{O_{ij} \in \mathcal{O}} \underline{C}(O_{ij},s_{t})$ over all operations can naturally be used as a measure of the makespan at step $t$. The reward $r_t$ is defined as the difference between the estimated makespan values at step $t$ and $t+1$  
\begin{equation}
r_t = \max \limits_{O_{ij} \in \mathcal{O}} \underline{C}(O_{ij},s_{t}) - \max \limits_{O_{ij} \in \mathcal{O}} \underline{C}(O_{ij},s_{t+1}).
\end{equation}

When the discounting factor $\gamma=1$, one can obtain the cumulative reward  $\sum_{t=0}^{|\mathcal{O}|-1}r_t = \max \limits_{O_{ij} \in \mathcal{O}} \underline{C}(O_{ij},s_{0}) - C_\text{max}$. In this way, maximizing the cumulative reward is equivalent to minimizing the makespan.

 \begin{figure}[t]
	\centering 
	\includegraphics[scale=0.54]{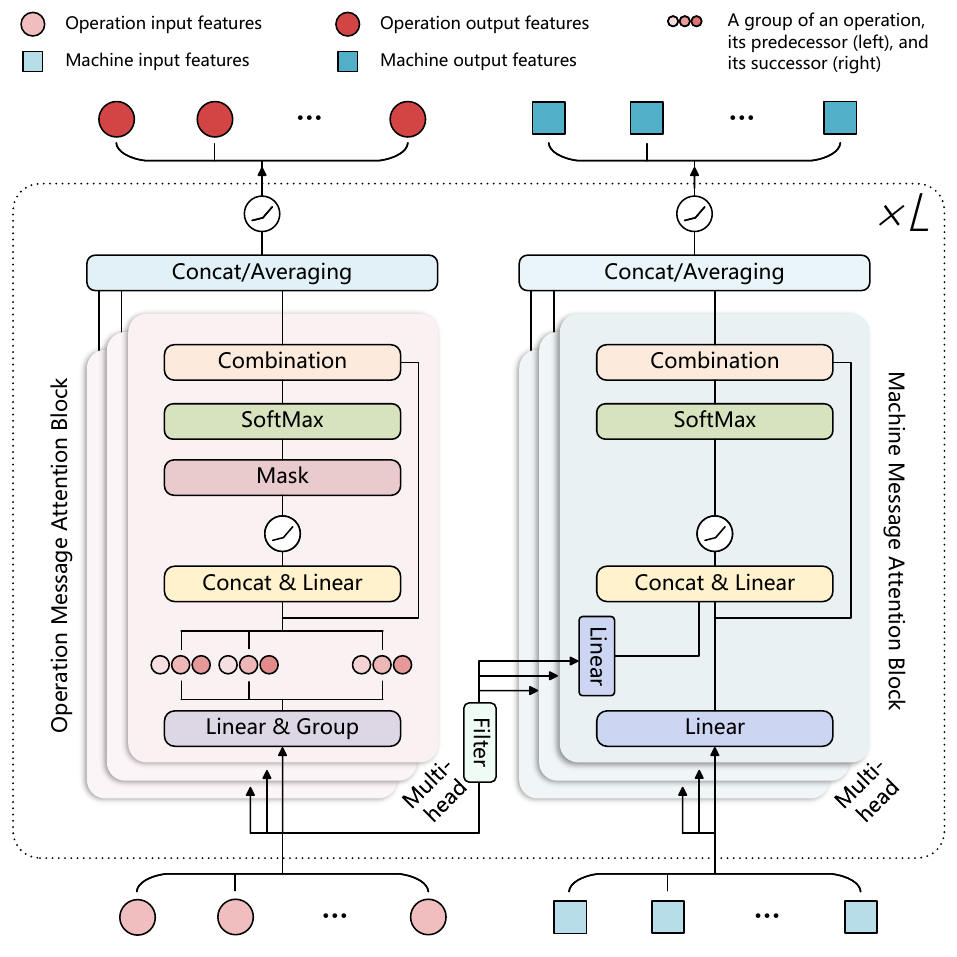}
	\caption{\small{The architecture of the dual attention network (DAN). 
	}}
	\label{fig:dan}
\end{figure}

\emph{Policy.} We employ a stochastic policy $\pi(a_t|s_t)$ whose distribution is generated by an DRL algorithm (an actor-critic network in our simulations) with trainable parameters $\Theta$. Given a state $s_t$, this distribution returns the probability of choosing each action $a_t \in \mathcal{A}(t)$.

\subsection{Dual attention network}

Self-attention models can relate elements of input sequences and find the significant ones by exploiting their relationships, which are suitable for finding high priority operations and/or machines. In addition, self-attention models can handle length-varying sequences (as inputs), which is well-tailored for solving FJSP instances across different scales. These considerations have motivated us to use self-attention models to extract features for operations and machines. The disjunctive graph of FJSP shows that there are multiple types of connections among operations, such as conjunctions and multiple groups of disjunctions. Each type can be seen as a kind of relationship. Therefore, it is  inappropriate to relate all operations through a uniform attention model. Moreover, the competitive relationship between machines is not explicitly explored in a disjunctive graph, but it is intuitive that they are crucial for modeling the priority of machines. 

To bypass these drawbacks, this paper proposes the dual attention network (DAN), an attention-based model for feature extraction of and between operations and machines in FJSP. The overall framework of DAN is shown in Fig. \ref{fig:dan}. DAN decomposes the complex relationships of operations and machines into two parts and handles them separately using two different attention blocks, which are called operation message attention block and machine message attention block, respectively. The former receives operation features as input and simply relates them through precedence constraints. The latter explicitly models the dynamic competitive relationships between machines, which handles the production state of each machine as well as its processing eligibility of the unscheduled operations. The connected two blocks altogether constitute a dual attention layer. Features of operations and machines are refined through $L$ layers in turn, which exhibit similar structures but with a different number of attention heads and parameters. Let $\omega$ collect all the parameters of DAN. Details for updating the $l$-th layer’s parameters at step $t$ are given as below. In the following, the subscripts $t$ and $l$ are omitted when clear from context.

\textit{1) Operation message attention block.} This block aims to relate operations in the same job so as to find the most significant operations via their inherent properties. Concretely, for each $O_{ij} \in \mathcal{O}_u$ with input features $h_{O_{ij}} \in \mathbb{R}^{d_{o}}$, this block models the relationships between $O_{ij}$, its predecessor $O_{i,j-1}$, and its successor $O_{i,j+1}$ (if it exists), by computing their attention coefficients as follows
\begin{equation}
	e_{i,j,p}=\text{LeakyReLU}\!\left(\vec{\mathbf{a}}^\top\!\left[(\mathbf{W} {h}_{O_{ij}}) \| (\mathbf{W} {h}_{O_{ip}})\right]\right)
\end{equation}
where $\vec{\mathbf{a}}^\top \in \mathbb{R}^{2d_{o}^{\prime}}$ and $\mathbf{W}\in \mathbb{R}^{d_{o}^{\prime} \times d_{o}}$ are linear transformations and for all $|p-j| \leq 1$.

Note that the computations are similar to those in GAT, but we have narrowed its scope. Due to the fact that the predecessor (or successor) of some operations may not exist or will be removed at some step, a dynamic mask on the attention coefficients of these predecessors and successors is performed. A softmax function is employed to normalize all $e_{i,j,p}$ across choices of $p$, obtaining normalized attention coefficients $\alpha_{i,j,p}$. Finally, the output feature vector ${h^{\prime}_{O_{ij}}} \in \mathbb{R}^{d_{o}^{\prime}}$ is obtained by a weighted linear combination of transformed input features $\mathbf{W} {h}_{O_{i,j-1}}$, $\mathbf{W} {h}_{O_{i,j}}$ and $\mathbf{W} {h}_{O_{i,j+1}}$ followed by a nonlinear activation function $\sigma$:
\begin{equation}
{h^{\prime}_{O_{ij}}}=\sigma\Big(\sum_{p =j-1}^{j+1} \alpha_{i,j,p} \mathbf{W} {h}_{O_{ip}}\Big)
\end{equation}

By connecting multiple operation message attention blocks one by one, the messages of $O_{ij}$ can be propagated to all operations in $J_i$.

\textit{2) Machine message attention block.} Two machines compete for the unscheduled operations that they can both process. This competitive relationship may change dynamically as production proceeds. We define $C_{kq}$ as the set of operations that machines $M_k$ and $M_q$ compete for. Let $\mathcal{N}_k =\{q \mid C_{kq} \ne \emptyset \}$ denote the set of competing machines with $M_k$. Additionally, we employ $c_{kq} = \sum_{O_{ij} \in C_{kq} \cap J_c} {h_{O_{ij}}}$ as a measure of the intensity of competition between $M_k$ and $M_q$, which makes sense as the competition becomes more severe when their competed candidates are more important. The machine message attention block leverages $c_{kq}$ to compute the attention coefficients $u_{kq}$ (The 'filter' operator in Fig. \ref{fig:dan} is used for obtaining $c_{kq}$ for each $M_k$ and $M_q$). For each $M_{k} \in \mathcal{M}_u$ with input features $h_{M_{k}} \in \mathbb{R}^{d_{m}}$, the attention coefficients $u_{kq}$ for all $q \in \mathcal{N}_k$ are computed as follows
\begin{equation}
	\label{mab1}
	u_{kq}=\text{LeakyReLU}\!\left(\vec{\mathbf{b}}^\top\!\left[(\mathbf{Z}^1 {h}_{M_{k}}) \| (\mathbf{Z}^1 {h}_{M_{q}}) \| (\mathbf{Z}^2 {c}_{{kq}})\right]\right)
\end{equation}
where $\mathbf{Z}^1 \in \mathbb{R}^{d_{m}^{\prime} \times d_{m}}$ and  $\mathbf{Z}^2 \in \mathbb{R}^{d_{m}^{\prime} \times d_{o}}$ are weight matrices and $\vec{\mathbf{b}}^\top \in \mathbb{R}^{3d_{m}^{\prime}}$ is a linear transformation.

Note that $C_{kk}$ is the set of unscheduled operations that $M_k$ can process and $k$ is always in $\mathcal{N}_k$. Therefore, \eqref{mab1} can be applied to compute $u_{kk}$, for $c_{kk}$ can be seen as a measure of the processing capacity of $M_k$. When $C_{kq} \cap J_c$ is empty, we fill $c_{kq}$ by zeros. Then, analogous softmax normalization  (across choices of $q$), combination, and activation steps are conducted to obtain the output features $h_{M_k}^{\prime} \in \mathbb{R}^{d_{m}^{\prime}}$.

\textit{3) Multi-head attention.} We apply multiple attention heads to both blocks for the purpose of learning a variety of relationships between entities. This technique has been shown  effective in strengthening the learning ability of attention models \cite{vaswani2017attention}. We specify its use in the operation message attention block, which is the same for the machine message attention block and thus omitted for brevity. Let $H$ be the number of attention heads in a dual attention layer and $H$ attention mechanisms with different parameters are applied. First, the computation of attention coefficients and the combination are performed in parallel. Then, their outputs are  fed into an aggregation operator for integration. Adopted from GAT, the aggregation operator refers to a concatenation except for the last layer which uses an averaging operator. Finally, the activation function $\sigma$ is applied to obtain the layer’s output.

\textit{4) Pooling.} Let the raw (handcrafted) features of operation $O_{ij}$ and machine $M_k$ denoted by $h_{O_{ij}}^{(0)}$  and $h_{M_{k}}^{(0)}$, respectively. Upon passing $h_{O_{ij}}^{(0)}$  and $h_{M_{k}}^{(0)}$ through $L$ dual attention layers, the learned features $h_{O_{ij}}^{(L)}$ and $h_{M_{k}}^{(L)}$ are ready for downstream decision-making tasks. Similar to \cite{kool2018attention}, we first apply an averaging pooling to the operation features and machine features separately, whose results are concatenated to form the global features of an FJSP instance; that is,
\begin{equation}
	h_{G}^{(L)} = \bigg[ \Big(\frac{1}{\left | \mathcal{O}_{u} \right | }\sum_{O_{ij}\in \mathcal{O}_{u}} h_{O_{ij}}^{(L)}\Big)
	\Big\| \Big(\frac{1}{\left | \mathcal{M}_{u} \right | }\sum_{M_{k}\in \mathcal{M}_{u}} h_{M_{k}}^{(L)}\Big) \bigg].
\end{equation}

\textit{5) Analysis.} Compared with the prior art \cite{lei2022multi,song2022flexible}, our method has advantages in the following aspects. First, we model the time-varying competition between machines explicitly use and rely on this relationship to construct machine features. On top of this, we employ the sum of processable candidate features to represent the intensity of competition and consider its influence on the machine's priority. Moreover, our design incurs a much smaller computational overhead, since the learning process can be understood as handling two (much) smaller graphs, one for operations and the other for machines. The graph for operations can be seen as a modified disjunctive graph with the node set $\mathcal{O}_u(t)$, where conjunctions are maintained but replaced by undirected edges and disjunctions are removed. The graph for machines is used for describing information incorporated in disjunctions, which takes  compatible machines as  nodes and characterizes their competitions by means of incorporating dynamic edge features. Such a representation features simplicity as well as intuition, for the complex connections have been grouped into two categories with $O(3   | \mathcal{O}  |  + { | \mathcal{M}  |}^2) $ connections in total. Since $ | \mathcal{O}  |$ is often much larger than $ | \mathcal{M}  |$  in practice, the number of connections is significantly reduced relative to existing disjunctive graph based results in \cite{lei2022multi,song2022flexible}. Last but not the least, irrelevant operations and machines get removed as scheduling proceeds, contributing to a decreasing state space and bringing computational efficiency.

\subsection{Decision-making module}
We design the decision-making network based on actor-critic RL which maintains the size-agnostic property of attention models. Two MLPs are used as the actor and critic, whose parameters are denoted by $\theta$ and $\phi$, respectively. The actor network aims to generate a stochastic policy $\pi_\theta(a_t|s_t)$ in two steps: it first produces a scalar $\mu(a_t|s_t)$ for each $a_t \in \mathcal{A}(t)$ and uses the softmax function to output the desired distribution. We concatenate all information related to $a_t = (O_{ij},M_k)$ (including the extracted features of $O_{ij}$ and $M_k$, the global features, and the compatible operation-machine pair features) in a single vector which is subsequently fed into $\text{MLP}_\theta$ to yield
\begin{equation}
	\mu(a_t|s_t)=\operatorname{MLP}_\theta\!\Big[ h_{O_{ij}}^{(L)} \big\|h_{M_k}^{(L)}  \big\|  h_{G}^{(L)} \big\|h_{(O_{ij},M_k)}  \Big].
\end{equation}
Then, the probability of choosing action $a_t$ is given by
\begin{equation}
	\pi_\theta\left(a_t \mid s_t\right)=\frac{\exp (\mu(a_t|s_t))}{\sum_{b_t \in \mathcal{A}(t)} \exp (\mu (b_t | s_t))}.
\end{equation}

The critic network is an estimator that takes the global features $h_{G}^{(L)}$ as input to produce a scalar $v_\phi(s_t)$, as an estimate of the state value. 

\begin{algorithm}[ht]
	\caption{Training DANIEL for FJSP}
	\label{alg:rdan}
	\begin{algorithmic}[1]
		\STATE {\bfseries Input:}
		A dual attention network with initial parameters $\Theta = \{\omega, \theta,\phi\}$, behavior actor network $\theta_{\rm old}$, pre-sampled training data $\mathcal{E}_{\rm  tr}$, and fixed validation data $\mathcal{E}_{\rm  val}$;
		\FOR {$n_{\rm ep} =1,2,...,N_{\rm ep}$}
		\STATE $\theta_{\rm old} \gets \theta$
		\FOR {$i =1,2,...,|\mathcal{E}_{\rm tr}|$}
		
		\FOR {$t =0, 1,..., T$}
		\STATE Sample $a_{i,t}$ using $\pi_{\theta_{\rm old}}(\cdot \mid s_{i,t})$;  
		\STATE Receive the reward $r_{i,t}$ and the new state $s_{i, t+1}$;
		\STATE Collect the transition $(s_{i,t},a_{i,t},r_{i,t},s_{i, t+1})$;
		\STATE Update $s_{i,t} \gets s_{i,t+1}$;
		\ENDFOR
		
		\STATE Compute the generalized advantage estimates $\hat{A}_{i,t}$ for $t =0, 1,..., T$ using collected transitions;

		\ENDFOR
		
		\FOR{$k = 1,2,...,K $}
		\STATE Compute the total loss $\sum_{i=1}^{|\mathcal{E}_{\rm tr}|}\ell_i^{\rm PPO}(\Theta)$; 
		\STATE Update all parameters $\Theta$;
		\ENDFOR

		\IF{\textit{Every} $N_{r}$ \textit{episodes}}
		\STATE Resample $|\mathcal{E}_{\rm tr}|$ instances to form the training data;
		\ENDIF
		
		\IF{\textit{Every} $N_{\rm val}$ \textit{episodes}}
		\STATE Validate $\pi_\theta$ on $\mathcal{E}_{\rm val}$;
		\ENDIF
		
		\ENDFOR
		\vspace{-0pt}
	\end{algorithmic}
\end{algorithm}

\subsection{Training procedure}
We employ the proximal policy optimization (PPO) algorithm \cite{schulman2017proximal} to train the proposed scheduling model. The generalized advantage estimation (GAE) technique \cite{schulman2015high} is utilized to stabilize training. The training procedure is summarized in Algorithm \ref{alg:rdan}. We train the model using $N_{\rm ep}$ episodes. For each episode \gang{$n_{\rm ep}$}, the agent interacts with a batch of same-scale FJSP environments $\mathcal{E}_{\rm tr}$ in parallel and collects the transition data, which are used for updating the model parameters $\Theta$. The environments are resampled every $N_{r}$ episodes, according to a fixed distribution. The policy is validated on some fixed validation data $\mathcal{E}_{\rm val}$ (with the same distribution as the training data) every $N_{\rm val}$ episodes. Two strategies for the action-selection are considered in our experiments. One is a greedy strategy that always chooses the action with the highest probability $\mu(a_t|s_t)$, used in validation. The other is an action-sampling strategy, i.e., sampling an action from the distribution $\pi_\theta$, used during training for sufficient exploration.

\section{Experiments}
\label{sec:experi}
We numerically compare the proposed DANIEL algorithm with several baselines including several popular PDRs, the exact method Google OR-Tools \footnote{https://developers.google.cn/optimization}, a genetic algorithm \cite{rooyani2019efficient}, and the DRL-based method \cite{song2022flexible} in this section. Both synthetically generated instances as well as popularly used FJSP benchmarks are used to verify its effectiveness in scheduling performance, generalization capability, and computational efficiency. 

\begin{table*}[ht!]
	\caption{Performance on two synthetic data of small to medium training size.}
	\label{tab:syn}
	\centering
	\footnotesize
	\setlength{\tabcolsep}{0.915em}
	\renewcommand{\arraystretch}{1.1}
	\begin{threeparttable}
		\begin{tabular}{ccc|cccc|cc|cc|c}
			
			\toprule[1pt]
			
			&       &       & \multicolumn{4}{c|}{PDRs}     & \multicolumn{2}{c|}{Greedy strategy} & \multicolumn{2}{c|}{Sampling strategy} &  \\
			& Size  &       & FIFO  & MORNR & SPT   & MWKR  & \cite{song2022flexible}  & DANIEL  & \cite{song2022flexible}  & DANIEL  & OR-Tools\tnote{1} \\
			
			\midrule
			\midrule
			
			\multirow{12}{*}{\rotatebox[origin=c]{90}{$\text{SD}_1$}}
			
			& \multirow{3}[1]{*}{$10 \times 5$} & Objective  & 119.40  & 115.38  & 129.82  & 113.23  & 111.67  & \textbf{106.71 } & 105.59  & \textbf{101.67 } & \multirow{3}[1]{*}{96.32 (5\%)} \\
			&       & Gap   & 24.06\% & 19.87\% & 34.76\% & 17.58\% & 16.03\% & \textbf{10.87\%} & 9.66\% & \textbf{5.57\%} &  \\
			&       & Time (s) & 0.16  & 0.16  & 0.16  & 0.16  & 0.45  & 0.45  & 1.11  & 0.74  &  \\
			
			\cmidrule{2-12}
			
			& \multirow{3}[0]{*}{$20 \times 5$} & Objective  & 216.08  & 214.16  & 230.48  & 209.78  & 211.22  & \textbf{197.56 } & 207.53  & \textbf{192.78 } & \multirow{3}[0]{*}{188.15 (0\%)} \\
			&       & Gap   & 14.87\% & 13.85\% & 22.56\% & 11.51\% & 12.27\% & \textbf{5.03\%} & 10.31\% & \textbf{2.46\%} &  \\
			&       & Time (s) & 0.32  & 0.32  & 0.32  & 0.32  & 0.90  & 0.94  & 2.36  & 1.87  &  \\
			
			\cmidrule{2-12}
			
			& \multirow{3}[0]{*}{$15 \times 10$} & Objective  & 184.55  & 173.15  & 198.33  & 171.25  & 166.92  & \textbf{161.28 } & 160.86  & \textbf{153.22 } & \multirow{3}[0]{*}{143.53 (7\%)} \\
			&       & Gap   & 28.65\% & 20.68\% & 38.22\% & 19.41\% & 16.33\% & \textbf{12.42\%} & 12.13\% & \textbf{6.79\%} &  \\
			&       & Time (s) & 0.51  & 0.51  & 0.50  & 0.50  & 1.43  & 1.35  & 3.98  & 3.89  &  \\
			
			\cmidrule{2-12}
			
			& \multirow{3}[1]{*}{$20 \times 10$} & Objective  & 233.48  & 219.80  & 255.17  & 216.11  & 215.78  & \textbf{198.50 } & 214.81  & \textbf{193.91 } & \multirow{3}[1]{*}{195.98 (0\%)} \\
			&       & Gap   & 19.22\% & 12.20\% & 30.25\% & 10.30\% & 10.15\% & \textbf{1.31\%} & 9.64\% & \textbf{-1.03\%} &  \\
			&       & Time (s) & 0.71  & 0.71  & 0.71  & 0.71  & 1.91  & 1.85  & 6.23  & 6.35  &  \\
			\midrule
			\midrule
			
			\multirow{12}{*}{\rotatebox[origin=c]{90}{$\text{SD}_2$}}
			
			& \multirow{3}[1]{*}{$10 \times 5$} & Objective  & 569.41  & 557.48  & 514.39  & 549.28  & 553.61  & \textbf{408.40 } & 483.90  & \textbf{366.74 } & \multirow{3}[1]{*}{326.24 (96\%)} \\
			&       & Gap   & 76.47\% & 72.52\% & 57.96\% & 70.01\% & 71.42\% & \textbf{25.68\%} & 49.71\% & \textbf{12.57\%} &  \\
			&       & Time (s) & 0.16  & 0.16  & 0.16  & 0.16  & 0.46  & 0.44 & 1.11  & 0.88  &  \\
			
			\cmidrule{2-12}
			
			& \multirow{3}[0]{*}{$20 \times 5$} & Objective  & 1045.83  & 1045.94  & 835.94  & 1026.03  & 1059.04  & \textbf{671.03 } & 962.90  & \textbf{629.94 } & \multirow{3}[0]{*}{602.04 (0\%)} \\
			&       & Gap   & 74.59\% & 74.58\% & 38.91\% & 71.31\% & 76.79\% & \textbf{11.52\%} & 60.70\% & \textbf{4.66\%} &  \\
			&       & Time (s) & 0.33  & 0.33  & 0.33  & 0.33  & 0.94  & 0.90  & 2.37  & 1.84  &  \\
			
			\cmidrule{2-12}
			
			& \multirow{3}[0]{*}{$15 \times 10$} & Objective  & 871.14  & 845.16  & 703.07  & 830.53  & 807.47  & \textbf{591.21 } & 756.07  & \textbf{521.83 } & \multirow{3}[0]{*}{377.17 (28\%)} \\
			&       & Gap   & 132.23\% & 125.32\% & 86.74\% & 121.45\% & 115.26\% & \textbf{57.16\%} & 101.52\% & \textbf{38.70\%} &  \\
			&       & Time (s) & 0.52  & 0.51  & 0.51  & 0.51  & 1.50  & 1.36  & 4.16  & 3.83  &  \\
			
			\cmidrule{2-12}
			
			& \multirow{3}[0]{*}{$20 \times 10$} & Objective  & 1088.05  & 1059.68  & 829.14  & 1040.69  & 1045.82  & \textbf{610.16 } & 990.37  & \textbf{552.64 } & \multirow{3}[0]{*}{464.16 (1\%)} \\
			&       & Gap   & 135.27\% & 129.09\% & 78.82\% & 124.98\% & 126.12\% & \textbf{31.58\%} & 114.15\% & \textbf{19.13\%} &  \\
			&       & Time (s) & 0.70  & 0.70  & 0.70  & 0.71  & 1.95  & 1.79  & 6.03  & 5.97  &  \\
			\bottomrule[1.5pt]
		\end{tabular}%
		
		\begin{tablenotes}
			\footnotesize
			\item[1] For OR-Tools, the solution and the ratio  of optimally solved instances are reported.
		\end{tablenotes}
		
	\end{threeparttable}
	
\end{table*}

\begin{table*}
	\caption{Performance on two synthetic data of larger-sizes.}
	\label{tab:large}
	\centering
	\footnotesize
	\setlength{\tabcolsep}{0.55em}
	\renewcommand{\arraystretch}{1.1}
	\begin{threeparttable}
		\begin{tabular}{ccc|cc|cccc|cccc|c}
			\toprule[1.5pt]
			&       &       & \multicolumn{2}{c|}{Top PDRs} & \multicolumn{4}{c|}{Greedy strategy}   & \multicolumn{4}{c|}{Sampling strategy} &  \\
			&       &       & \multirow{2}[1]{*}{SPT} & \multirow{2}[1]{*}{MWKR} & \cite{song2022flexible}  & \cite{song2022flexible}  & DANIEL  & DANIEL  & \cite{song2022flexible}  & \cite{song2022flexible}  & DANIEL  & DANIEL  &  \\
			& Size  &       &       &       & $10 \times 5$  & $20 \times 10$ & $10 \times 5$  & $20 \times 10$ & $10 \times 5$  & $20 \times 10$ & $10 \times 5$  & $20 \times 10$ & OR-Tools\tnote{1}  \\
			\midrule
			\midrule
			\multirow{6}{*}{\rotatebox[origin=c]{90}{$\text{SD}_1$}}
			
			& \multicolumn{1}{c}{\multirow{3}[2]{*}{$30 \times 10$}} & Objective  & 350.07  & 312.93  & 314.71  & 313.04  & 288.61  & \textbf{281.49 } & 308.55  & 312.59  & 286.77  & \textbf{279.20 } & \multirow{3}[2]{*}{274.67 (6\%)} \\
			&       & Gap   & 27.47\% & 13.96\% & 14.61\% & 14.01\% & 5.10\% & \textbf{2.50\%} & 12.36\% & 13.49\% & 4.43\% & \textbf{1.67\%} &  \\
			&       & Time (s) & 1.09  & 1.09  & 2.86  & 2.84  & 2.78  & 2.76  & 12.79  & 12.80  & 12.37  & 12.37  &  \\
			\cmidrule{2-14}      & \multirow{3}[2]{*}{$40 \times 10$} & Objective  & 445.17  & 414.82  & 417.87  & 416.18  & 379.28  & \textbf{371.45 } & 410.76  & 415.25  & 379.71  & \textbf{370.48 } & \multirow{3}[2]{*}{365.96 (3\%)} \\
			&       & Gap   & 21.66\% & 13.37\% & 14.21\% & 13.75\% & 3.65\% & \textbf{1.52\%} & 12.26\% & 13.49\% & 3.77\% & \textbf{1.14\%} &  \\
			&       & Time (s) & 1.50  & 1.50  & 3.82  & 3.81  & 3.77  & 3.77  & 24.54  & 24.40  & 22.58  & 21.38  &  \\
			\midrule
			\midrule
			\multirow{6}{*}{\rotatebox[origin=c]{90}{$\text{SD}_2$}}
			
			& \multirow{3}[2]{*}{ $30 \times 10$} & Objective  & 1105.99  & 1539.67  & 1564.57  & 1543.69  & 794.62  & \textbf{774.56 } & 1486.56  & 1461.16  & 757.48  & \textbf{725.27 } & \multirow{3}[2]{*}{692.26 (0\%)} \\
			&       & Gap   & 59.74\% & 122.89\% & 126.55\% & 123.57\% & 14.85\% & \textbf{11.95\%} & 115.21\% & 111.51\% & 9.47\% & \textbf{4.80\%} &  \\
			&       & Time (s) & 1.10  & 1.10  & 2.93  & 2.93  & 2.80  & 2.75  & 12.88  & 12.75  & 12.48  & 12.17  &  \\
			\cmidrule{2-14}      & \multirow{3}[2]{*}{$40 \times 10$} & Objective  & 1357.16  & 2037.65  & 2048.96  & 2032.54  & 983.37  & \textbf{962.58 } & 1976.25  & 1945.53  & 951.21  & \textbf{914.02 } & \multirow{3}[2]{*}{998.39 (0\%)} \\
			&       & Gap   & 38.74\% & 108.66\% & 109.87\% & 108.12\% & 0.52\% & \textbf{-1.67\%} & 102.45\% & 99.26\% & -2.74\% & \textbf{-6.60\%} &  \\
			&       & Time (s) & 1.50  & 1.51  & 3.87  & 3.93  & 3.76  & 3.74  & 24.55  & 24.50  & 21.45  & 21.09  &  \\
			\bottomrule[1pt]
			
		\end{tabular}%
		
		\begin{tablenotes}
			\footnotesize
			\item[1] For OR-Tools, the solution and the ratio  of optimally solved instances are reported. 
		\end{tablenotes}
		
	\end{threeparttable}

\end{table*}

\subsection{Datasets}
An FJSP instance with $n$ jobs and $m$ machines is denoted as ``$n \times m$'' in short hereafter (the number of operations varies in different instances). We consider two types of synthetic data with distinct distributions to examine the learning as well as generalization performance of the proposed DANIEL algorithm. 

\gang{The first is adapted from \cite{song2022flexible}, which allows jobs to have a varying number of operations, denoted by $\text{SD}_1$ in the Table.} We refer interested readers to the original text \cite{song2022flexible} for details. The second is a task with a wider range for the random processing time of operations, denoted by $\text{SD}_2$ in the Table below. Specifically, for an $n \times m$ FJSP instance generated for $\text{SD}_2$, each job has $m$ operations.  For each $O_{ij}$, the values $|\mathcal{M}_{ij}|$ and $p_{ij}^{k}$ are integers sampled from $U(1,m)$ and $U(1,99)$, respectively. For a fair comparison with the approach in \cite{song2022flexible}, we consider six different scales of FJSP instances using the two data: $10 \times 5$, $20 \times 5$, $15 \times 10$, $20 \times 10$, $30 \times 10$, and $40 \times 10$. We train our model on four smaller sizes of each data with randomly generated instances ($8$ models in total), while testing is performed on two extra larger ones in addition to those. Both the testing data (unseen) and the validation data are generated in advance each having $100$ instances. Furthermore, we evaluate the trained models on four groups of public benchmarks to explore their capabilities on cross-distribution tasks, including  \emph{mk1-10} instances in \cite{brandimarte1993routing} and $3$ groups of \emph{la1-40} instances (except sdata) in \cite{hurink1994tabu}.

\begin{table*}[t]
	\caption{Performance on public benchmarks.}
	\label{tab:bench}
	\centering
	\footnotesize
	\setlength{\tabcolsep}{0.65em}
	\renewcommand{\arraystretch}{1.15}
	\begin{threeparttable}
		\begin{tabular}{cc|c|cccc|cccc|cc}
			\toprule[1pt]
			&       &  \multicolumn{1}{c|}{Top PDR}      & \multicolumn{4}{c|}{Greedy-stategy}   & \multicolumn{4}{c|}{Sampling-stategy}   &       &  \\
			&       &       & \cite{song2022flexible}  & \cite{song2022flexible}  & DANIEL  & DANIEL  & \cite{song2022flexible}  & \cite{song2022flexible}  & DANIEL  & DANIEL  &       &  \\
			\multicolumn{1}{l}{Benchmark} &       & MWKR  & $10 \times 5$  & $15 \times 10$ & $10 \times 5$  & $15 \times 10$ & $10 \times 5$  & $15 \times 10$ & $10 \times 5$  & $15 \times 10$ & 2SGA  & OR-Tools \\
			\midrule
			\midrule
			\multirow{3}[2]{*}{mk} & Objective  & 201.74  & 201.40  & 198.50  & 185.70  & \textbf{184.40 } & 190.30  & 190.60  & \textbf{180.80 } & 180.90  & 175.20  & 174.20  \\
			& Gap   & 28.91\% & 28.52\% & 26.77\% & 13.58\% & \textbf{12.97\%} & 18.56\% & 19.0\% & 9.53\% & \textbf{8.95\%} & 3.17\% & 1.5\% \\
			& Time (s) & 0.49  & 1.26  & 1.25  & 1.29  & 1.30  & 4.13  & 4.13  & 4.12  & 4.08  & 57.60  & 1447.08  \\
			\midrule
			\multirow{3}[2]{*}{la (rdata)} & Objective  & 1053.10  & 1030.83  & \textbf{1031.33 } & 1031.63  & 1040.05  & 985.30  & 988.38  & \textbf{978.28 } & 983.33  & \multirow{3}[2]{*}{-} & 935.80  \\
			& Gap   & 13.86\% & 11.15\% & \textbf{11.14\%} & 11.42\% & 12.07\% & 5.57\% & 5.95\% & \textbf{4.95\%} & 5.49\% &       & 0.11\% \\
			& Time (s) & 0.52  & 1.40  & 1.40  & 1.37  & 1.36  & 4.81  & 4.82  & 4.71  & 4.73  &       & 1397.43  \\
			\midrule
			\multirow{3}[2]{*}{la (edata)} & Objective  & 1219.01  & 1187.48  & 1182.08  & 1194.98  & \textbf{1175.53 } & \textbf{1116.68 } & 1119.43  & 1122.60  & 1119.73  & \multirow{3}[2]{*}{-} & 1028.93  \\
			& Gap   & 18.6\% & 15.53\% & 15.0\% & 16.33\% & \textbf{14.41\%} & \textbf{8.17\%} & 8.69\% & 9.08\% & 8.72\% &       & -0.03\% \\
			& Time (s) & 0.52  & 1.40  & 1.40  & 1.37  & 1.38  & 4.91  & 4.87  & 4.73  & 4.70  &       & 899.60  \\
			\midrule
			\multirow{3}[2]{*}{la (vdata)} & Objective  & 952.01  & 955.90  & 954.33  & \textbf{944.85 } & 948.73  & 930.80  & 931.33  & \textbf{925.40 } & 925.68  & 812.20\tnote{1}  & 919.60  \\
			& Gap   & 4.22\% & 4.25\% & 4.02\% & \textbf{3.28\%} & 3.75\% & 1.32\% & 1.34\% & \textbf{0.69\%} & 0.72\% & 0.39\% & -0.01\% \\
			& Time (s) & 0.52  & 1.37  & 1.37  & 1.37  & 1.37  & 4.71  & 4.72  & 4.77  & 4.75  & 51.43  & 639.17  \\
			\bottomrule[1pt]
		\end{tabular}%
		
		\begin{tablenotes}
			\footnotesize
			\item[1] The makespan and gap of 2SGA on la(vdata) benchmark are computed on la1-30 instances, as reported in \cite{rooyani2019efficient}. 
		\end{tablenotes}
		
	\end{threeparttable}
\end{table*}%

\subsection{Configurations}
The training configurations were kept the same as in \cite{song2022flexible} for comparison, with $N_{\rm ep}=1,000$, $|\mathcal{E}_{\rm tr}|$, $N_{r}=20$, and $N_{\rm val}=10$. Hyperparameters of algorithms were tuned on $\text{SD}_2$ instances of size $10 \times 5$ and kept the same for all $8$ models.  For a trade-off between performance and computational efficiency, \gang{our DAN used $L=2$ dual attention layers and each block used $H=4$ attention heads per layer with ELU as activation function  $\sigma$.} The output dimensions of each attention head in the two blocks are $d_o^{(1)}=d_m^{(1)}=32$ for the first layer and $d_o^{(2)}=d_m^{(2)}=8$ for the second layer. Both $\text{MLP}_\theta$ and $\text{MLP}_\phi$ have two hidden layers of dimension $64$ with tanh as  activation. For the PPO parameters, \gang{the policy, value, and entropy coefficient in the loss funciton} were set to be $1$, $0.5$, and $0.01$. The clipping parameter $\epsilon$, GAE parameter $\lambda$, and discounting factor $\gamma$ were set to be $0.2$, $0.98$, and $1$, respectively. During training, we updated the network for $K=4$ epochs per episode via the Adam optimizer \gang{\cite{kingma2014adam}} with the learning rate $lr = 3 \times 10^{-4}$. All experiments were carried out on a machine with an Intel Xeon Platinum $8163$ CPU and a single NVIDIA Tesla T4 GPU. \gang{The code is available. \footnote{https://github.com/wrqccc/FJSP-DRL}}

\subsection{Baselines and performance metrics}
First, \gang{we selected four PDRs for operation sequencing that have been shown to yield good performance in  \cite{sels2012comparison}}, and generalized them to solve FJSP. In the experiments, we reported the results averaged over $5$ independent runs for each PDR due to their stochastic nature. Specifics regarding the implementation are stated as follows.
\begin{itemize}
	\item First in first out (FIFO): selecting the earliest ready candidate operation and the earliest ready compatible machine.
	\item Most operations remaining (MOPNR): selecting the candidate operation with the most remaining successors and a machine which can immediately process it.
	\item Shortest processing time (SPT): selecting the compatible operation-machine pair with the shortest processing time.
	\item Most work remaining (MWKR): selecting the candidate operation with the most remaining successor average processing time and a machine which can immediately process it.
\end{itemize}

Second, the results from OR-Tools and an advanced genetic algorithm  were chosen as a reference line for time-consuming but high-performance methods. As an authoritative constrained programming solver, OR-Tools was employed to generate (near-)optimal solutions with $1,800$ seconds set as the time limit. In \cite{rooyani2019efficient}, a two-stage genetic algorithm (2SGA) was designed to solve FJSP with improved performance and efficiency than regular genetic algorithms, whose results on two groups of benchmarks were directly imported 
due to lack of open-source implementations. Last but not the least, we compared DANIEL with the state-of-the-art DRL method recently proposed in \cite{song2022flexible}. We used their open-source code \footnote{https://github.com/songwenas12/fjsp-drl} to perform training and testing following their default settings. We directly imported their results on $\text{SD}_1$, including the trained models and validation instances of four smaller sizes as well as testing instances of all six sizes. We reported the results of DANIEL and DRL for both greedy and sampling-based action-selection strategies in testing as done in   \cite{song2022flexible}. Specifically, the sampling strategy uses action-sampling to solve an instance for $100$ times in parallel and records the best one, to improve the solution quality at the price of an acceptable computation burden. Model’s performance was evaluated in terms of the average makespan as well as the relative gap between its makespan and the best-known solution, which is either the solution of OR-Tools for synthetic instances or the best result reported in \cite{behnke2012test} for public benchmarks. In our experiments,  performance of the two strategies was compared separately and the best results for the DRL-based methods and PDRs are emphasized in bold for each problem.

\begin{figure}[h]
	\includegraphics[scale=0.49]{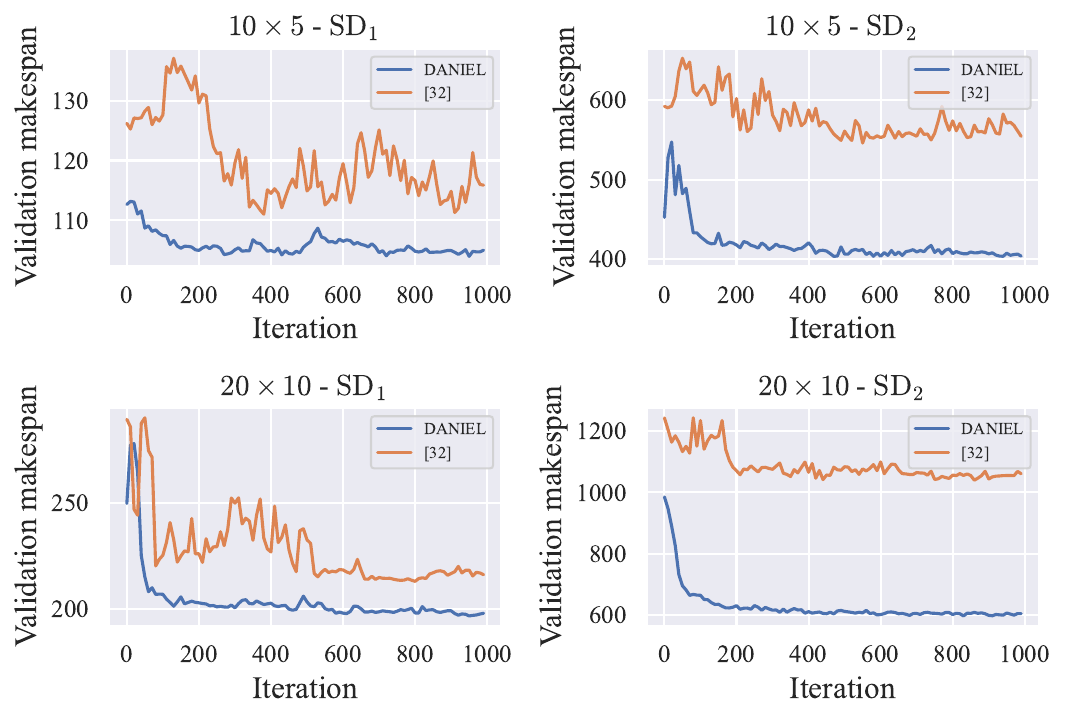}
	\caption{\small{\gang{Training curve of the DRL \cite{song2022flexible} and DANIEL algorithms on $10 \times 5$ and $20 \times 10$ FJSP instances using SD$_1$ and SD$_2$. }
	}}
	\label{fig:traincurve}
\end{figure}

\subsection{Results on synthetic data}
In Table \ref{tab:syn}, we reported each model's average makespan and gap on testing instances which are generated from the same scales using the same distribution as its training instances. It is clear from the bold numbers that, for both data across all problem sizes, DANIEL not only outperforms all PDRs by a significant margin, but also exhibits a notable improvement relative to the DRL solution in \cite{song2022flexible} for both action-selection strategies. Moreover, the optimality gap of DANIEL is less than $5\%$ in three tasks when using the sampling strategy.  Most excitingly, DANIEL even beats OR-Tools by $1.03\%$ on the $20 \times 10$ FJSP instances on $\text{SD}_1$ data, \gang{while all instances in this data weren't solved optimally by OR-Tools within the time limits}. The advantage of DANIEL becomes more pronounced on  $\text{SD}_2$ data. When the processing time range gets larger, performance of both PDRs and the DRL \cite{song2022flexible} are affected, resulting in a considerable gap to the best solution. In this case, DANIEL still performs well, especially when using the sampling strategy. By means of the proposed tight state representation, DANIEL has runtime comparable to PDRs and is computationally efficient in general. \gang{Its runtime is close to that of \cite{song2022flexible}, although implementations of the simulation environment of FJSP in the two algorithms are different.} Another interesting phenomenon from the results of PDRs is that SPT consistently performs the worst for all problem sizes on $\text{SD}_1$ data but performs the best for all sizes on $\text{SD}_2$, demonstrating its unstable performance over varying tasks. This may be caused by the very different processing time ranges, since a compatible operation-machine pair with the shortest processing time may have a high priority on $\text{SD}_2$ but a low priority on $\text{SD}_1$ data.

The training curves of DANIEL and DRL \cite{song2022flexible} on $10 \times 5$ and $20 \times 10$ FJSP instances using SD$_1$ and SD$_2$ are presented in Fig. \ref{fig:traincurve}, where the averaged makespan over $100$ validation instances are shown. It is worth mentioning that our method and \cite{song2022flexible}  used the same training data, validation frequency, and validation data. It can be observed that our method converges to a better solution more smoothly for both data, corroborating its powerful and stable performance. 

Furthermore, we examined the generalization performance of the models trained over $10 \times 5$ and $20 \times 10$ tasks on two data using $30 \times 10$ and $40 \times 10$ testing instances, respectively. DANIEL was compared with \cite{song2022flexible} and the best PDRs in $\text{SD}_1$ and $\text{SD}_2$. As shown in Table \ref{tab:large}, DANIEL consistently generates high-quality solutions for large-scale problems with acceptable computation cost, which are considerably better than all baseline methods and occasionally beat OR-Tools. \gang{Especially for the $40 \times 10$ FJSP instances on $\text{SD}_1$}, the model trained on $20 \times 10$ instances with the sampling strategy outperforms OR-Tools by $6.60\%$. These results show that DANIEL can learn general knowledge by training on small-scale tasks, which can be employed to solve unseen large-scale instances. Note that the model trained on $20 \times 10$ consistently outperforms that trained on $10 \times 5$ instances, which is expected because its size is closer to that of the testing data.

\subsection{Results on benchmarks}
Cross-distribution application is critical for a model in practical use, as real-world problems may come from unknown distributions. Therefore, we further explore the performance of models (trained on synthetic data) on four groups public benchmarks whose distributions are completely different from the training instances. Each benchmark comprises instances having different problem sizes. For example, the number of jobs and machines in the mk benchmark ranges from $10$ to $20$ and from $5$ to $15$, respectively. We selected the models trained on $10 \times 5$ and $15 \times 10$ using $\text{SD}_1$ data for testing, which achieved the best results on these benchmarks in \cite{song2022flexible}. The results of MWKR (the best among the four PDRs), the OR-Tools, the GA baseline in \cite{rooyani2019efficient}, and the models of \cite{song2022flexible} (with greedy and sampling strategies) trained on the same data are presented in Table \ref{tab:bench}. It can be seen that both OR-Tools and 2SGA are generally far ahead in performance but take quite a long time for computations. In contrast, the two DRL methods achieve good performance which is better than the best PDR with acceptable runtime. Again, DANIEL surpasses \cite{song2022flexible} in most cases and exhibits comparable performance in the remaining ones (the testing of greedy strategy on la (rdata) and sampling strategy on la (edata), where the difference is less than 0.6\%). Particularly, DANIEL outperforms \cite{song2022flexible} by a big margin for both action-selection strategies on mk instances. These results showcase that DANIEL can genuinely capture the inherent structural information of FJSP and discriminate compatible pairs with high priority rather than only learning the regularity behind a specific data distribution. We believe that DANIEL can perform better on real-world problems with \gang{unknown distributions} upon a small amount of fine-tuning, as DANIEL can outperform  the exact methods on tasks with known distributions, which is left for future research. 

\section{Conclusions}
\label{sec:concl}
This study proposes DANIEL, a novel end-to-end learning framework for addressing standard FJSPs. DANIEL combines the attention mechanism and DRL while maintaining the size-agnostic property, i.e., training on small-size problems and deploying to larges-size ones. DANIEL builds on a concise representation of the complex FJSP structure, by employing two attention blocks to explore relationships between operations within the job as well as between the competing machines. The dual attention network receives abundant information about decision-relevant operations and machines as input and extracts their features through GAT-style attention mechanisms. The downstream decision-making network generates composite decisions for the operation sequencing and machine assignment problems simultaneously using actor-critic reinforcement learning, trained by the PPO algorithm. Substantial numerical results on synthetic data as well as publicly available benchmarks show that DANIEL performs significantly better than traditional PDRs and the state-of-the-art DRL method while being computationally efficient. In several cases, DANIEL even beats the OR-Tools, which has never been reported for end-to-end learning approaches in the literature. Moreover, DANIEL demonstrates excellent generalization performance to large-scale and real-world instances when trained using small-size FJSP data relative to existing DRL methods. \gang{In the future, extending DANIEL to address dynamic FJSP with uncertainty, such as new job insertions, and to the multi-agent setting constitutes interesting topics for future research.}

\section{Appendix}
\label{sec:Apdix}
Here,  details for the three types of feature vectors at state $s_t$ ($t$ is omitted below) are provided.

\textit{1) Features of operations:} 4 static attributes and 6 dynamic properties are recorded for each $O_{ij}\in \mathcal{O}_u$:
\begin{itemize}
	\item Minimum processing time among all machines.
	\item Average processing time among all machines.
	\item Span of processing time among all machines.
	\item Proportion of machines that $O_{ij}$ can be processed on.
	\item Scheduling tag: the value is $0$ if $O_{ij}$ is unscheduled otherwise $1$.
	\item Estimated Lower bound of the completion time:  $\underline{C}(O_{ij})$ defined in the text.
	\item Job remaining number of operations: the number of unscheduled operations in $J_i$.
	\item Job remaining workload: the sum of average processing time of unscheduled operations in $J_i$.
	\item Waiting time: the time from the ready time until $T_s$. (0 for $O_{ij}$ that is not ready yet)
	\item Remaining processing time: the time from $T_s$ until the completion time. (0 for unscheduled $O_{ij}$)
\end{itemize}

\textit{2) Features of machines:} Each machine $M_k \in \mathcal{M}_u$ owns a feature vector with 8 elements and the first two are static:
\begin{itemize}
	\item Minimum processing time among all operations.
	\item Average processing time of operations that $M_k$ can process.
	\item Number of unscheduled operations that $M_k$ can process.
	\item Number of candidates that $M_k$ can process.
	\item Free time: the moment when $M_k$ is free.
	\item Waiting time: the time from the free time until $T_s$. (0 for working $M_k$)
	\item Working tag: the value is $0$ if $M_{k}$ is free otherwise $1$.
	\item Remaining processing time: the time from $T_s$ until the free time. (0 for free $M_k$)
\end{itemize}

\textit{3) Features of compatible operation-machine pairs:} When a compatible pair $(O_{ij},M_{k})$ is considered, we record 8 features for it and the first two are static: 
\begin{itemize}
	\item Processing time $p_{ij}^{k}$.
	\item Ratio of $p_{ij}^{k}$ to the maximum processing time of $O_{ij}$.
	\item Ratio of $p_{ij}^{k}$ to the maximum processing time of candidates that $M_k$ can process.
	\item Ratio of $p_{ij}^{k}$ to the maximum processing time of unscheduled operations.
	\item Ratio of $p_{ij}^{k}$ to the maximum processing time of unscheduled operations that $M_k$ can process.
	\item Ratio of $p_{ij}^{k}$ to the maximum processing time of compatible pairs.
	\item Ratio of $p_{ij}^{k}$ to remaining workload of $J_i$.
	\item Summation of waiting time of $O_{ij}$ and $M_k$.
\end{itemize} 

\bibliographystyle{IEEEtran}
\bibliography{DANIEL}

\begin{thebibliography}{10}
\providecommand{\url}[1]{#1}
\csname url@samestyle\endcsname
\providecommand{\newblock}{\relax}
\providecommand{\bibinfo}[2]{#2}
\providecommand{\BIBentrySTDinterwordspacing}{\spaceskip=0pt\relax}
\providecommand{\BIBentryALTinterwordstretchfactor}{4}
\providecommand{\BIBentryALTinterwordspacing}{\spaceskip=\fontdimen2\font plus
\BIBentryALTinterwordstretchfactor\fontdimen3\font minus
  \fontdimen4\font\relax}
\providecommand{\BIBforeignlanguage}[2]{{%
\expandafter\ifx\csname l@#1\endcsname\relax
\typeout{** WARNING: IEEEtran.bst: No hyphenation pattern has been}%
\typeout{** loaded for the language `#1'. Using the pattern for}%
\typeout{** the default language instead.}%
\else
\language=\csname l@#1\endcsname
\fi
#2}}
\providecommand{\BIBdecl}{\relax}
\BIBdecl

\bibitem{mao2019opportunities}
S.~Mao, B.~Wang, Y.~Tang, and F.~Qian, ``Opportunities and challenges of
  artificial intelligence for green manufacturing in the process industry,''
  \emph{Eng.}, vol.~5, no.~6, pp. 995--1002, Dec. 2019.

\bibitem{eng2022}
J.~{Chen}, J.~{Sun}, and G.~{Wang}, ``From unmanned systems to autonomous
  intelligent systems,'' \emph{Eng.}, vol.~12, no.~5, pp. 16--19, 2022.

\bibitem{ding2019defining}
K.~Ding, F.~T. Chan, X.~Zhang, G.~Zhou, and F.~Zhang, ``Defining a digital
  twin-based cyber-physical production system for autonomous manufacturing in
  smart shop floors,'' \emph{Int. J. Prod. Res.}, vol.~57, no.~20, pp.
  6315--6334, Jan. 2019.

\bibitem{arunarani2019task}
A.~Arunarani, D.~Manjula, and V.~Sugumaran, ``Task scheduling techniques in
  cloud computing: {A} literature survey,'' \emph{Future Gener. Comput. Syst.},
  vol.~91, pp. 407--415, Feb. 2019.

\bibitem{satunin2014multi}
S.~Satunin and E.~Babkin, ``A multi-agent approach to intelligent
  transportation systems modeling with combinatorial auctions,'' \emph{Expert
  Syst. Appl.}, vol.~41, no.~15, pp. 6622--6633, Nov. 2014.

\bibitem{zhang2019review}
J.~Zhang, G.~Ding, Y.~Zou, S.~Qin, and J.~Fu, ``Review of job shop scheduling
  research and its new perspectives under {I}ndustry 4.0,'' \emph{J. Intell.
  Manuf.}, vol.~30, pp. 1809--1830, 2019.

\bibitem{xie2019review}
J.~Xie, L.~Gao, K.~Peng, X.~Li, and H.~Li, ``Review on flexible job shop
  scheduling,'' \emph{IET Collab. Intell. Manuf.}, vol.~1, no.~3, pp. 67--77,
  Sep. 2019.

\bibitem{meng2020mixed}
L.~Meng, C.~Zhang, Y.~Ren, B.~Zhang, and C.~Lv, ``Mixed-integer linear
  programming and constraint programming formulations for solving distributed
  flexible job shop scheduling problem,'' \emph{Comput. Ind. Eng.}, vol. 142,
  p. 106347, Apr. 2020.

\bibitem{demir2013evaluation}
Y.~Demir and S.~K. {\.I}{\c{s}}leyen, ``Evaluation of mathematical models for
  flexible job-shop scheduling problems,'' \emph{Appl. Math. Modell.}, vol.~37,
  no.~3, pp. 977--988, Feb. 2013.

\bibitem{mnih2013playing}
V.~Mnih, K.~Kavukcuoglu, D.~Silver, A.~A. Rusu, J.~Veness, M.~G. Bellemare,
  A.~Graves, M.~Riedmiller, A.~K. Fidjeland, G.~Ostrovski \emph{et~al.},
  ``Human-level control through deep reinforcement learning,'' \emph{Nature},
  vol. 518, no. 7540, pp. 529--533, Feb. 2015.

\bibitem{wang2019genetic}
M.~Wang and B.~Xin, ``A genetic algorithm for solving flexible flow shop
  scheduling problem with autonomous guided vehicles,'' in \emph{IEEE Intl.
  Conf. Control Autom.}\hskip 1em plus 0.5em minus 0.4em\relax IEEE, 2019, pp.
  922--927.

\bibitem{rooyani2019efficient}
D.~Rooyani and F.~M. Defersha, ``An efficient two-stage genetic algorithm for
  flexible job-shop scheduling,'' \emph{IFAC-PapersOnLine}, vol.~52, no.~13,
  pp. 2519--2524, 2019.

\bibitem{zhang2009effective}
G.~Zhang, X.~Shao, P.~Li, and L.~Gao, ``An effective hybrid particle swarm
  optimization algorithm for multi-objective flexible job-shop scheduling
  problem,'' \emph{Comput. Ind. Eng.}, vol.~56, no.~4, pp. 1309--1318, May
  2009.

\bibitem{du2018high}
W.~Du, W.~Zhong, Y.~Tang, W.~Du, and Y.~Jin, ``High-dimensional robust
  multi-objective optimization for order scheduling: {A} decision variable
  classification approach,'' \emph{IEEE Trans. Ind. Inf.}, vol.~15, no.~1, pp.
  293--304, May 2018.

\bibitem{li2019hybrid}
J.-Q. Li, M.-X. Song, L.~Wang, P.-Y. Duan, Y.-Y. Han, H.-Y. Sang, and Q.-K.
  Pan, ``Hybrid artificial bee colony algorithm for a parallel batching
  distributed flow-shop problem with deteriorating jobs,'' \emph{IEEE Trans.
  Cybern.}, vol.~50, no.~6, pp. 2425--2439, Oct. 2019.

\bibitem{haupt1989survey}
R.~Haupt, ``A survey of priority rule-based scheduling,'' \emph{Oper. Res.
  Spektrum}, vol.~11, no.~1, pp. 3--16, Mar. 1989.

\bibitem{wang2021review}
L.~Wang, Z.~Pan, and J.~Wang, ``A review of reinforcement learning based
  intelligent optimization for manufacturing scheduling,'' \emph{Compl. Syst.
  Model. Simul.}, vol.~1, no.~4, pp. 257--270, Dec. 2021.

\bibitem{wang2021dynamic}
L.~Wang, X.~Hu, Y.~Wang, S.~Xu, S.~Ma, K.~Yang, Z.~Liu, and W.~Wang, ``Dynamic
  job-shop scheduling in smart manufacturing using deep reinforcement
  learning,'' \emph{Comput. Netw.}, vol. 190, p. 107969, May 2021.

\bibitem{luo2021real}
S.~Luo, L.~Zhang, and Y.~Fan, ``Real-time scheduling for dynamic
  partial-no-wait multiobjective flexible job shop by deep reinforcement
  learning,'' \emph{IEEE Trans. Autom. Sci. Eng.}, vol.~19, no.~4, pp.
  3020--3038, Aug. 2021.

\bibitem{du2022reinforcement}
Y.~Du, J.~Li, C.~Li, and P.~Duan, ``A reinforcement learning approach for
  flexible job shop scheduling problem with crane transportation and setup
  times,'' \emph{IEEE Trans. Neural Netw. Learn. Syst.}, Oct. 2022, DOI:
  10.1109/TNNLS.2022.3208942.

\bibitem{brandimarte1993routing}
P.~Brandimarte, ``Routing and scheduling in a flexible job shop by tabu
  search,'' \emph{Ann. Oper. Res.}, vol.~41, no.~3, pp. 157--183, Sep. 1993.

\bibitem{vaswani2017attention}
A.~Vaswani, N.~Shazeer, N.~Parmar, J.~Uszkoreit, L.~Jones, A.~N. Gomez,
  {\L}.~Kaiser, and I.~Polosukhin, ``Attention is all you need,'' \emph{Adv.
  Neural Inf. Process. Syst.}, vol.~30, 2017.

\bibitem{nazari2018reinforcement}
M.~Nazari, A.~Oroojlooy, L.~Snyder, and M.~Tak{\'a}c, ``Reinforcement learning
  for solving the vehicle routing problem,'' \emph{Adv. Neural Inf. Process.
  Syst.}, vol.~31, 2018.

\bibitem{kool2018attention}
H.~V.~H. W.~Kool and M.~Welling, ``Attention, learn to solve routing
  problems!'' \emph{Proc. Int. Conf. Learn. Represent.}, 2018.

\bibitem{mirhoseini2021graph}
A.~Mirhoseini, A.~Goldie, M.~Yazgan, J.~W. Jiang, E.~Songhori, S.~Wang, Y.-J.
  Lee, E.~Johnson, O.~Pathak, A.~Nazi \emph{et~al.}, ``A graph placement
  methodology for fast chip design,'' \emph{Nature}, vol. 594, no. 7862, pp.
  207--212, June 2021.

\bibitem{kwon2021matrix}
Y.-D. Kwon, J.~Choo, I.~Yoon, M.~Park, D.~Park, and Y.~Gwon, ``Matrix encoding
  networks for neural combinatorial optimization,'' \emph{Adv. Neural Inf.
  Process. Syst.}, vol.~34, pp. 5138--5149, 2021.

\bibitem{manchanda2020gcomb}
S.~Manchanda, A.~Mittal, A.~Dhawan, S.~Medya, S.~Ranu, and A.~Singh, ``{GCOMB:
  L}earning budget-constrained combinatorial algorithms over billion-sized
  graphs,'' \emph{Adv. Neural Inf. Process. Syst.}, vol.~33, pp.
  20\,000--20\,011, 2020.

\bibitem{zhang2020learning}
C.~Zhang, W.~Song, Z.~Cao, J.~Zhang, P.~S. Tan, and X.~Chi, ``Learning to
  dispatch for job shop scheduling via deep reinforcement learning,''
  \emph{Adv. Neural Inf. Process. Syst.}, vol.~33, pp. 1621--1632, 2020.

\bibitem{park2021learning}
J.~Park, J.~Chun, S.~H. Kim, Y.~Kim, and J.~Park, ``Learning to schedule
  job-shop problems: {R}epresentation and policy learning using graph neural
  network and reinforcement learning,'' \emph{Int. J. Prod. Res.}, vol.~59,
  no.~11, pp. 3360--3377, 2021.

\bibitem{chen2022deep}
R.~Chen, W.~Li, and H.~Yang, ``A deep reinforcement learning framework based on
  an attention mechanism and disjunctive graph embedding for the job-shop
  scheduling problem,'' \emph{IEEE Trans. Ind. Inf.}, vol.~19, no.~2, pp.
  1322--1331, Apr. 2022.

\bibitem{lei2022multi}
K.~Lei, P.~Guo, W.~Zhao, Y.~Wang, L.~Qian, X.~Meng, and L.~Tang, ``A
  multi-action deep reinforcement learning framework for flexible job-shop
  scheduling problem,'' \emph{Expert Syst. Appl.}, vol. 205, p. 117796, Nov.
  2022.

\bibitem{song2022flexible}
W.~Song, X.~Chen, Q.~Li, and Z.~Cao, ``Flexible job-shop scheduling via graph
  neural network and deep reinforcement learning,'' \emph{IEEE Trans. Industr.
  Inform.}, vol.~19, no.~2, pp. 1600--1610, Feb. 2023.

\bibitem{huang2019convolutional}
G.~Huang, Z.~Liu, G.~Pleiss, L.~Van Der~Maaten, and K.~Q. Weinberger,
  ``Convolutional networks with dense connectivity,'' \emph{IEEE Trans. Pattern
  Anal. Mach. Intell.}, vol.~44, no.~12, pp. 8704--8716, May 2019.

\bibitem{velivckovic2017graph}
P.~Veli{\v{c}}kovi{\'c}, G.~Cucurull, A.~Casanova, A.~Romero, P.~Lio, and
  Y.~Bengio, ``Graph attention networks,'' \emph{Proc. Int. Conf. Learn.
  Represent.}, 2017.

\bibitem{schulman2017proximal}
J.~Schulman, F.~Wolski, P.~Dhariwal, A.~Radford, and O.~Klimov, ``Proximal
  policy optimization algorithms,'' \emph{arXiv:1707.06347}, 2017.

\bibitem{schulman2015high}
J.~Schulman, P.~Moritz, S.~Levine, M.~Jordan, and P.~Abbeel, ``High-dimensional
  continuous control using generalized advantage estimation,''
  \emph{arXiv:1506.02438}, 2015.

\bibitem{hurink1994tabu}
J.~Hurink, B.~Jurisch, and M.~Thole, ``Tabu search for the job-shop scheduling
  problem with multi-purpose machines,'' \emph{Oper. Res. Spektrum}, vol.~15,
  pp. 205--215, Dec. 1994.

\bibitem{kingma2014adam}
D.~P. Kingma and J.~Ba, ``Adam: {A} method for stochastic optimization,''
  \emph{arXiv:1412.6980}, 2014.

\bibitem{sels2012comparison}
V.~Sels, N.~Gheysen, and M.~Vanhoucke, ``A comparison of priority rules for the
  job shop scheduling problem under different flow time-and tardiness-related
  objective functions,'' \emph{Int. J. Prod. Res.}, vol.~50, no.~15, pp.
  4255--4270, Aug. 2012.

\bibitem{behnke2012test}
D.~Behnke and M.~J. Geiger, ``Test instances for the flexible job shop
  scheduling problem with work centers,'' \emph{Arbeitspapier/Research
  Paper/Helmut-Schmidt-Universit{\"a}t, Lehrstuhl f{\"u}r
  Betriebswirtschaftslehre, insbes. Logistik-Management}, 2012.

\end{thebibliography}

	\end{document}